\documentclass[letterpaper, 10 pt, conference]{ieeeconf}
\IEEEoverridecommandlockouts
\usepackage{cite}
\usepackage{amsmath,amssymb,amsfonts}
\usepackage{algorithmic}
\usepackage[xetex]{graphicx}
\usepackage{textcomp}
\usepackage{todonotes}
\usepackage{xcolor}
\def\BibTeX{{\rm B\kern-.05em{\sc i\kern-.025em b}\kern-.08em
    T\kern-.1667em\lower.7ex\hbox{E}\kern-.125emX}}

\usepackage[acronym, toc]{glossaries}
\usepackage{siunitx}
\usepackage[pdftex,bookmarks=true]{hyperref}
\usepackage[pdftex]{hyperref}
\usepackage[noabbrev]{cleveref}
\usepackage{subcaption}
\usepackage{booktabs}

\newacronym{drl}{DRL}{deep reinforcement learning}
\newacronym{rl}{RL}{reinforcement learning}
\newacronym{mdp}{MDP}{Markov decision process}
\newacronym{ann}{ANN}{artificial neural network}
\newacronym{ddpg}{DDPG}{Deep Deterministic Policy Gradient}
\newacronym{ros}{ROS}{Robot Operating System}
\newacronym{her}{HER}{Hindsight Experience Replay}
\newacronym{sgd}{SGD}{Stochastic Gradient Descent}
\newacronym{shap}{SHAP}{SHapley Additive exPlanations}
\newacronym{xai}{XAI}{explainable artificial intelligence}
\newacronym{dof}{DOFs}{degrees-of-freedom}
\newacronym{dnn}{DNN}{deep neural network}
\newacronym{ai}{AI}{artificial intelligence}
\newacronym{lime}{LIME}{Local Interpretable Model-agnostic Explanations}
\newacronym{ml}{ML}{machine learning}

\newcommand\thefont{\expandafter\string\the\font}

\begin{document}
\title{\LARGE \bf
Causal versus Marginal Shapley Values for Robotic Lever Manipulation Controlled using Deep Reinforcement Learning
}

\author{Sindre Benjamin Remman$^{1}$, Inga Strümke$^{2}$ and Anastasios M. Lekkas$^{3}$
\thanks{$^{1}$Sindre Benjamin Remman is with the Department of Engineering Cybernetics, 
        Norwegian University of Science and Technology (NTNU), Trondheim, Norway
        {\tt\small sindre.b.remman@ntnu.no}}%
\thanks{$^{2}$Inga Strümke is with the Department of Engineering Cybernetics, 
        Norwegian University of Science and Technology (NTNU), Trondheim, Norway
        {\tt\small inga.strumke@ntnu.no}}%
\thanks{$^{3}$Anastasios M. Lekkas is with the Department of Engineering Cybernetics,
        Centre for Autonomous Marine Operations and Systems (AMOS),
        Norwegian University of Science and Technology (NTNU), Trondheim, Norway
        {\tt\small anastasios.lekkas@ntnu.no}}%
}

\maketitle

\begin{abstract}
We investigate the effect of including domain knowledge about a robotic system's causal relations when generating explanations. To this end, we compare two methods from explainable artificial intelligence, the popular KernelSHAP and the recent causal SHAP, on a deep neural network trained using deep reinforcement learning on the task of controlling a lever using a robotic manipulator. A primary disadvantage of KernelSHAP is that its explanations represent only the features' direct effects on a model's output, not considering the indirect effects a feature can have on the output by affecting other features. Causal SHAP uses a partial causal ordering to alter KernelSHAP's sampling procedure to incorporate these indirect effects. This partial causal ordering defines the causal relations between the features, and we specify this using domain knowledge about the lever control task. We show that enabling an explanation method to account for indirect effects and incorporating some domain knowledge can lead to explanations that better agree with human intuition. This is especially favorable for a real-world robotics task, where there is considerable causality at play, and in addition, the required domain knowledge is often handily available. 
\end{abstract}
\begin{keywords}
Deep reinforcement learning, robotics, explainable artificial intelligence, Shapley additive explanations, causal SHAP
\end{keywords}

\section{Introduction}

Data-driven control methods have become widespread over the last years due to their ability to capture changes in system dynamics and adapt accordingly.
The control system can use such methods to successfully adapt to situations that the engineer cannot envision beforehand. 
Reinforcement learning-based methods have shown great promise in terms of adaptability in robotics applications. However, this adaptability comes at a cost, as \gls{rl} methods are often paired with a function approximator such as a \gls{dnn}, which are in general not interpretable for humans. The combination of \gls{rl} with \glspl{dnn} is called \gls{drl} and has been prominent in the last decade because of its high performance on several difficult tasks \cite{AlphaZero,badia_agent57_2020}. \gls{drl} has also had success within robotic manipulation \cite{levine2015endtoend, kalashnikov_qt-opt_2018, gu_deep_2016}. However, the non-interpretable nature of \glspl{dnn} implies that using \gls{drl} to control a real cyber-physical system during safety-critical operation is not prudent. 

The problem with interpretability permeates the current \gls{ml} state-of-the-art. Based on this, scientists are currently researching how to explain the decisions of \gls{ml} agents, or even how to make the agents explain themselves. The field addressing these issues is called \gls{xai}, from which a steadily increasing number of methods are being developed. Among the first and most widely used \gls{xai} methods is \gls{lime}, presented in \cite{ribeiro2016whyshoulditrustyou_lime}. This method locally approximates the uninterpretable model using an interpretable model, for instance, a linear model. 
Linear \gls{lime} in an instance of an \emph{additive feature attribution methods}, whose defining property is having an explanation model that is a linear function of binary variables. This was formalized by \cite{Lundberg2017}, who also realized that several explanation methods share this property, thus unifying several explanation methods and introducing \gls{shap}. The \gls{shap} framework produces feature attributions satisfying the axioms of the Shapley decomposition, a solution concept from cooperative game theory~\cite{shapley_value_1952}. Adapting linear LIME to satisfy the Shapley axioms results in the feature attribution method KernelSHAP~\cite{Lundberg2017}.

In the case of \gls{shap}, the binary variables indicate presence or absence of a model feature, as the Shapley value is calculated by considering all possible arrangements of contributors to an outcome. \gls{shap} implementations, therefore, rely on calculating an \gls{ml} model's expected outcome in the absence of model features. In KernelSHAP~\cite{Lundberg2017}, this sampling is done using a marginal distribution of the excluded features, which amounts to assuming independence between the model features. As argued by~\cite{heskes_causal_2020}, explanations generated using the marginal distribution can only represent the direct effects of features on the model, not the indirect effects~\cite{pearl_direct_indirect}. Taking the causal structure in the data into account, \cite{heskes_causal_2020} present a modification to the \gls{shap} package, named \textit{causal} \gls{shap}. 


The contributions of this paper are the following:

\begin{itemize}
    \item We employ causal \gls{shap} for explaining a system that involves robotic manipulation. To achieve this, we analyze the system to uncover its causal structure.
    \item We compare the explanations generated by KernelSHAP and causal \gls{shap}. In doing so, we investigate the effect of taking indirect feature effects into account, thereby obtaining feature attributions based on a more complete physical description of the system at hand.
\end{itemize}
This paper is organized as follows: in~\Cref{sec:preliminaries}, we present the necessary theory behind \gls{drl}, \gls{shap} and causal \gls{shap}; in~\Cref{sec:methods}, we describe the task to be solved using \gls{drl}, the experimental design, and how we use \gls{shap} and causal \gls{shap} to explain the decision-making agent; in~\Cref{sec:results_discussion}, we present and discuss results obtained; and finally, in~\Cref{sec:conclusion}, we draw our conclusions. 

\section{Preliminaries}\label{sec:preliminaries}
This section gives an overview of the theory and terminology necessary to understand the remainder of this paper. Firstly, we give an overview of the fundamentals of \gls{drl}. Secondly, the theory behind \gls{shap} is explained. Lastly, we look at how \gls{shap} is modified to create causal \gls{shap} values.

\subsection{Deep Reinforcement Learning}
\begin{figure}
    \centering
    \vspace{0.15cm}
    \includegraphics[width=\linewidth]{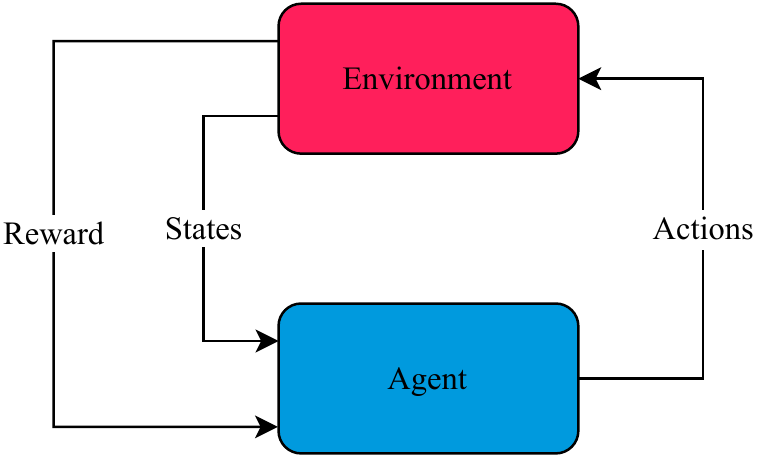}
    \caption{The reinforcement learning loop.}
    \label{fig:RL_loop}
\end{figure}

In \gls{rl}, we divide the system into two parts: the \emph{agent} and the \emph{environment}. The interactions between the two are illustrated in~\Cref{fig:RL_loop}. The agent receives a state from the environment, performs an action based on this state, and receives a new state together with a reward from the environment. This cycle then repeats for the whole operation. The goal of \gls{rl} is to find a \emph{policy} that maps states to actions. The so-called \emph{optimal policy} does this in an optimal way, in the sense that it maximizes the long-term expected reward, defined by the discounted infinite horizon model: 
\setlength{\textfloatsep}{2pt}
\begin{equation}
    E[\sum^\infty_{t=0}\gamma^t r_t] \,,
\end{equation}
\setlength{\textfloatsep}{2pt}
where $\gamma \in [0,1]$ is the discount factor, and $r_t$ is the reward received at time $t$ \cite[pp.13-15]{ReinfLearnState}.

As previously stated, \gls{drl} refers to \gls{rl} where the function approximator is a \gls{dnn}. Here, we use a \gls{drl} agent trained using the \gls{ddpg} \cite{DDPGpaper} algorithm. This is an actor-critic algorithm, which means that it trains two neural networks, the \emph{actor-network} and the \emph{critic-network}. The actor-network functions as the policy, which means that it maps states to actions, and the critic-network is used to guide the training of the actor-network. From a control engineering point of view, the policy is then akin to a controller. \gls{ddpg} trains a deterministic policy, which means that a specific policy will always give the same output for the same input. 

\subsection{Shapley Additive Explanations}
The Shapley decomposition, introduced by Lloyd Shapley in 1953 \cite{shapley_value_1952}, has in recent years been applied extensively in the \gls{xai} literature. It is a solution concept from cooperative game theory and distributes a game's outcome among the participants while uniquely preserving efficiency, monotonicity, and equal treatment, see, for instance, Theorem 2 in \cite{Young:1985aa}. The Shapley value of participant $i$ is calculated as a weighted mean over all subsets $\mathcal{S}\subseteq\mathcal{N}$ of the game's $N$ participants, not containing participant $i$:
\begin{equation}
    \phi_i = \sum_{\mathcal{S} \subseteq \mathcal{N}} \frac{|\mathcal{S}|!\left(N-|\mathcal{S}|-1\right)!}{N!} 
    \left(v(\mathcal{S} \cup \{j\})- v(\mathcal{S}) \right) \,.
    \label{eq:shapley}
\end{equation}
Here, $v(\mathcal{S})$ is the \textit{characteristic function}, which maps any set of participants in the game to a single real number $2^{\mathcal{N}}\to \mathbb{R}$, and thus fully characterizes the game.

In the context of \gls{xai}, the game can, for instance, be represented by a \gls{ml} model performing a prediction task, the result of the game by the model prediction, and the participants of the game by the model's input features. We can then obtain a feature attribution from the Shapley decomposition, quantifying how each of the input features affects the model prediction. The prediction of a \gls{ml} model $f$ trained on a set of data with features $\mathbf{x}$, made on the specific input features $\mathbf{x}^{\ast}$, can be decomposed as follows
\begin{equation}
    f(\mathbf{x}^{\ast}) = \phi_0 + \sum_{i=1}^N \phi^{\ast}_i \,,
\end{equation}
with $\phi_0$ the expected value of the model output across the data set, $E[f(\textbf{x})]$, and $\phi^{\ast}_i$ the Shapley value for the specific prediction on $\mathbf{x} = \mathbf{x}^{\ast}$.

There are two main challenges associated with calculating Shapley values for feature attribution: First, the calculation is very computationally expensive. A model using $N$ features would need to be evaluated $2^N$ times, once for each feature's inclusion or exclusion, as is readily seen from~\Cref{eq:shapley}. Second, it is in general not possible to evaluate a fitted \gls{ml} model with sets of features missing. To circumvent these challenges, implementations such as the widely used \gls{shap}, introduced by Lundberg and Lee \cite{Lundberg2017}, rely on approximations. The \gls{shap} calculation uses as characteristic function
an estimate of the expected model prediction, conditional upon the values of the included features, $\mathbf{x}_\mathcal{S} = \mathbf{x}^{\ast}_\mathcal{S}$, namely
\begin{equation}
    v(\mathcal{S}) = E[f(\mathbf{x})| \mathbf{x}_\mathcal{S} 
    = \mathbf{x}^{\ast}_{\mathcal{S}}] \,,
    \label{eq:v_SHAP}
\end{equation}
in the notation of~\cite{aas_explaining_2020}.
As such, \gls{shap} values attribute the change in the expected model prediction to each model feature by estimating how much each feature contributes to driving the model prediction away from its mean prediction across a data set. 
KernelSHAP, introduced in~\cite{Lundberg2017}, estimates~\Cref{eq:v_SHAP} with absent features using a marginal distribution, which amounts to the assumption of independence between the included and excluded features.

Various changes to the sampling procedure used for estimating the expected model prediction have since been suggested, and particularly relevant in this context are \cite{aas_explaining_2020,frye2020,janzing2020,heskes_causal_2020}.


\subsection{Causal SHAP}\label{sec:causal_shap}
In order to calculate the expected model prediction, Heskes et al.\ suggest \cite{heskes_causal_2020} adapting the sampling procedure in the \gls{shap} calculation, conditioning the absent features upon the values of the included features by \textit{intervention}. 
First suggested by Aas et al.\cite{aas_explaining_2020}, the \gls{shap} calculation can use the conditional distribution of the excluded features, instead of the marginal. Furthermore, interventional probabilities can be inferred from conditional probabilities using Pearl's $do$-calculus~\cite{pearl_do,pearl_do_revisited}. Combining conditional \gls{shap} with the $do$-calculus thus allows us to use the interventional distribution in the \gls{shap} calculation. Then, \Cref{eq:v_SHAP} becomes
\begin{equation}
    v(\mathcal{S}) = E[f(\mathbf{x})|do(\mathbf{x}_{\mathcal{S}}=\mathbf{x}_{\mathcal{S}}^{\ast})] \,,
\end{equation}
which we calculate by integrating over the absent features $\overline{\mathcal{S}}$, as detailed in~\cite{heskes_causal_2020}. 
The main advantage of the resulting so-called \textit{causal} \gls{shap} values is that both direct as well as \textit{indirect} effects of the model features are taken into account. The direct effects represent the change in the model's prediction due to a change in a feature without changing the absent features. The indirect effects, on the other hand, represent the change caused in the absent features by the intervention upon a feature, see equations (5) in~\cite{heskes_causal_2020}.
The inclusion of these indirect effects constitutes the main difference between causal \gls{shap} values and the marginal \gls{shap} values introduced earlier, as the latter by construction only represent indirect effects.

The causal SHAP implementation used in this paper\footnote{\url{https://github.com/sbremman/causal_shap_python}} was created by the first author, by adapting the \texttt{R} implementation by \cite{heskes_causal_2020} to \texttt{Python}. In the implementation, we specify the causal structure of the data via a (partial) causal ordering in the form of a nested list in which each list is defined as causally dependent on the elements in the preceding list(s). In addition, we use a second, separate list to specify whether the dependencies within each nested list result from a confounding factor or mutual interactions between the components in the nested list.

\section{Methodology}\label{sec:methods}

In this section, we describe the task of the \gls{drl} agent, its training, our dataset creation, and finally, how we analyze it using the two different \gls{shap} implementations.

\subsection{Lever manipulation task}
\begin{figure}
    \centering
    \vspace{0.15cm}
    \includegraphics[trim=0 11cm 0 15cm,clip,width=.75\linewidth ]{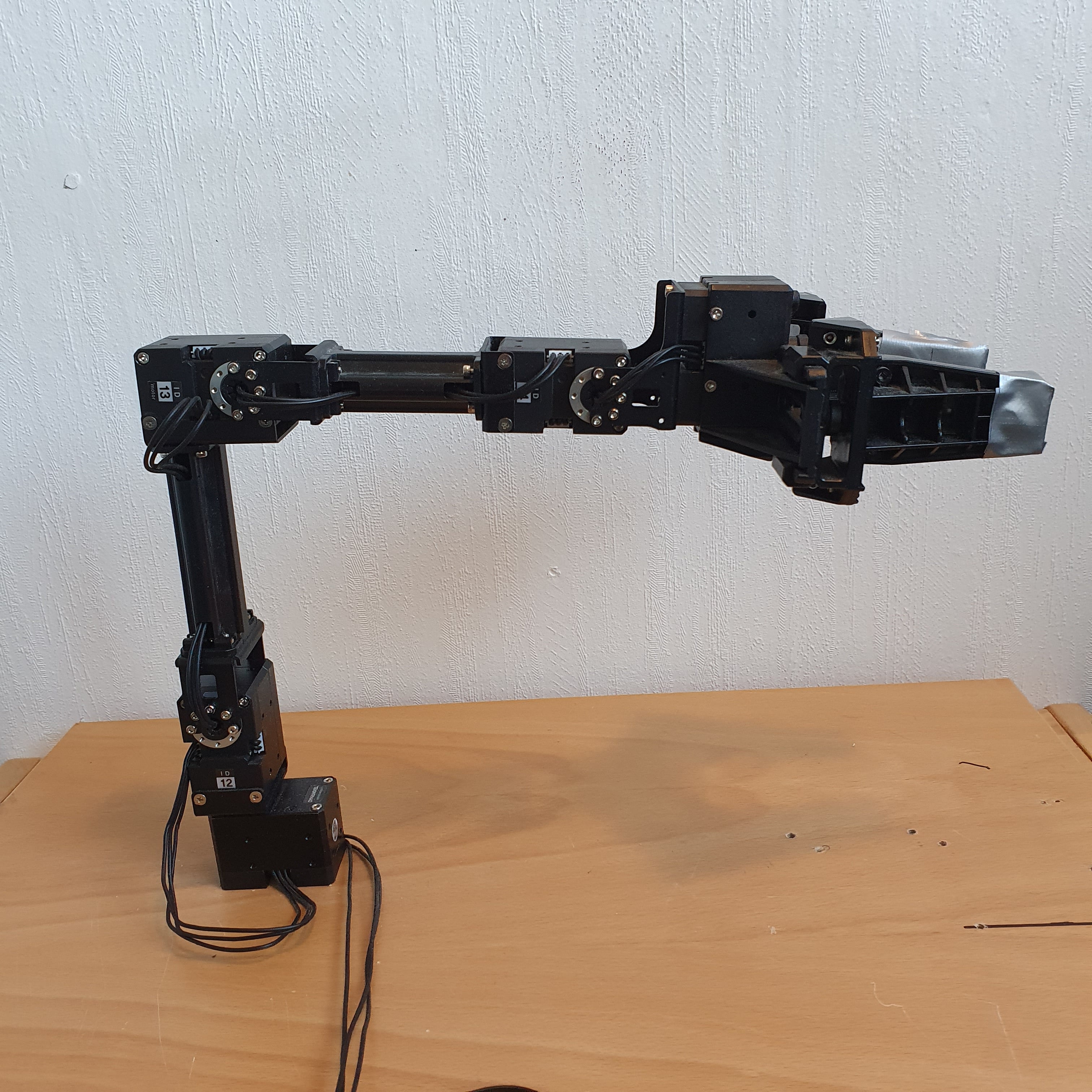}
    \caption{The OpenMANIPULATOR-X.}
    \label{fig:open_manipulator}
\end{figure}

The robotic manipulator that is used in this paper is the OpenMANIPULATOR-X\footnote{\url{https://emanual.robotis.com/docs/en/platform/openmanipulator_x/overview/}} by Robotis, which can be seen in \Cref{fig:open_manipulator}. 
This manipulator has five degrees of freedom, four for the joints and one for the gripper. We do not use the first joint during this lever manipulation task, which corresponds to a rotation about the manipulator's base. This is both because this makes the training more efficient, but also because the angle of this joint is trivial to solve for using $$\theta_1 = \text{arctan2}(y_{lever},x_{lever}),$$ where $y_{lever}$ and $x_{lever}$ is the $y$- and $x$- coordinates expressed in the inertial frame of the manipulator. 

The task involves moving the lever from a randomly selected start angle to a randomly selected target angle. These target and start angles are selected uniformly according to $$\theta_{start},\theta_{target} \in \mathbb{R}: \theta_{start},\theta_{target} \in [-1.0\text{ } \si{\radian}, 1.0\text{ } \si{\radian}],$$ and
$|\theta_{start}-\theta_{target}|>0.4  \text{ }\si{\radian}.$

\subsection{States and Actions}
\begin{figure}
    \centering
    \vspace{0.15cm}
    \includegraphics[width=\linewidth]{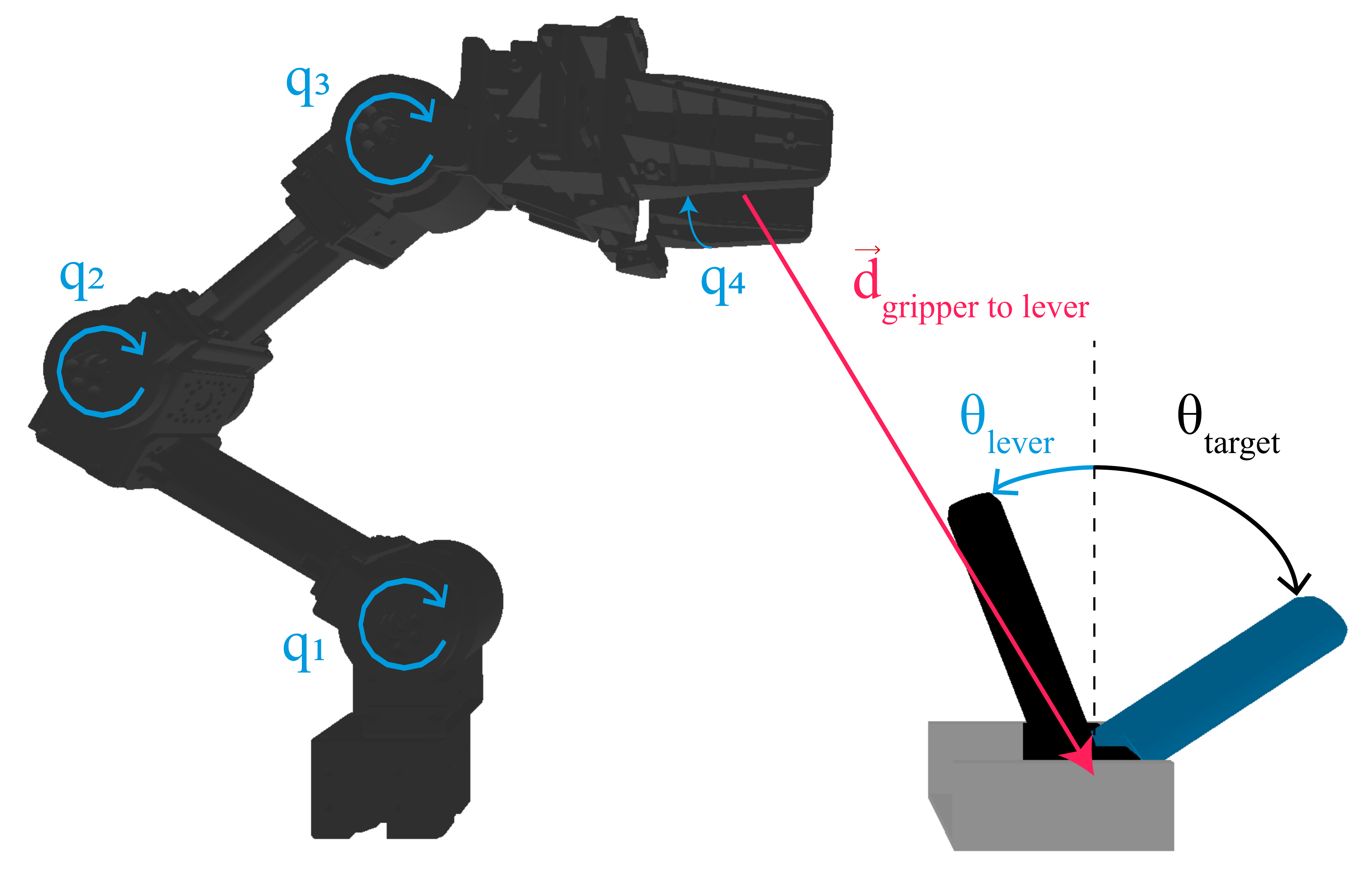}
    \caption{Visualization of the system's states.}
    \label{fig:manip_and_lever_states}
\end{figure}

In this task, the dimension of the state-space is eight and consists of the joint angles of the manipulator, the distance between the two fingers of the gripper, the horizontal and vertical distance from the end-effector to the lever, and the current and desired angles of the lever. See \Cref{fig:manip_and_lever_states} below for a visualization of the system's states.

The action vector is of dimension four, where the first three entries correspond to the desired relative movement of the shoulder, elbow and wrist joints respectively. The fourth entry in the action-vector indicates whether the gripper should open or close: 
\begin{equation*}
\begin{split}
a_4 \geq 0, & \rightarrow \text{Gripper should open} \\
a_4 < 0, & \rightarrow \text{Gripper should close.}
\end{split}
\end{equation*}

\subsection{Training procedure}
The agent was trained using the \gls{ddpg} algorithm together with the technique \gls{her} \cite{andrychowicz2017hindsight_her}. \gls{her} enables the usage of sparse rewards, and we, therefore, give sparse rewards according to 
$$r =  \begin{cases}
    -1, & \text{if } |\theta_{lever}-\theta_{target}|\geq0.025 \text{ }\si{\radian}\\
    0, & \text{if } |\theta_{lever}-\theta_{target}|<0.025 \text{ }\si{\radian}
\end{cases},$$
where $0.025$ rad is the chosen precision for the lever manipulation task. The agent is trained in simulations, and we use two different simulators for this. The first simulator is PyBullet, which is a fast simulator, but for our purposes, not close enough to the real-world environment. The agent trained in PyBullet is then transfer learned in Gazebo, which is slower, but more similar to the real-world environment. After transfer learning in Gazebo, we can deploy the agent in the real-world environment. This training procedure is described in more detail in \cite{remman_robotic_2021}, where the main difference here is that we have reduced the number of input features, which was done primarily to make the feature attributions simpler to interpret.

\subsection{Dataset and explanations}\label{sec:methodology_xai}
In order to sample the excluded features, all implementations of \gls{shap} rely on background datasets. Furthermore, we wish to choose interesting decisions by our agent to explain. To this end, we collect a dataset by letting the fully trained \gls{drl} agent operate in the real-world environment by running $15$ \emph{test episodes} with randomly selected target and start lever angles. From these test episodes, we identify a collection of interesting events and compare the explanations generated by the two different \gls{shap} implementations on these. We remove the episodes where these events occur and use the resulting dataset as our background data set. We chose one event from episode $1$ and two events from episode $3$, meaning that our background dataset consists of episodes $2$ and $4-15$. 

To display the results, we create a plot similar to the force plot available in the \gls{shap} package\footnote{\url{https://github.com/slundberg/shap}}. 
The force plot illustrates the ``force'' of each feature on the prediction, showing how the features force the prediction away from the mean prediction and towards the model's prediction. 
In our adaptation, we show one force plot for each of the agent's actions on the same figure. This is primarily done to compactly convey the information and compare the different actions for the same decision.

\begin{figure}
    \centering
    \vspace{0.15cm}
    \includegraphics{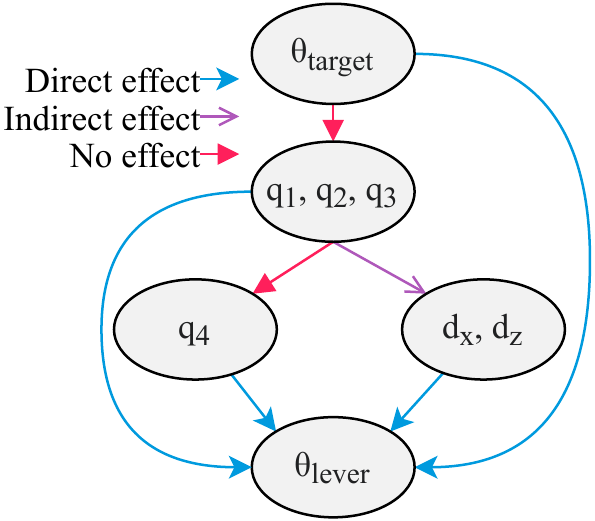}
    \caption{Visualization of the chosen causal ordering.}
    \label{fig:causal_graph}
\end{figure}

As stated in \Cref{sec:causal_shap}, the causal SHAP implementation requires the causal structure of the data to be specified by a causal ordering. The causal ordering we use is: [[$\theta_{target}$], [$q_1, q_2, q_3$], [$q_4, d_x, d_z$], [$\theta_{lever}$]]. In addition, we assume that none of the features are influenced by a confounding factor but instead have only mutual interactions. A visualization of our causal ordering is shown in \Cref{fig:causal_graph}, where blue arrows indicate each feature's direct effect on the target, purple arrows indirect effects via other features, and the red arrows indirect effects that we know are not present. This is described below.

In our setup, $\theta_{target}$ reveals a weakness in the causal \gls{shap} implementation: This feature does not have any causal connection with any of the other features, but because of the way causal \gls{shap} is implemented, it still has to be defined in the causal ordering. We choose to list this feature first in the causal ordering to avoid the indirect effects of all the other features flowing through this independent feature (which would happen if we placed it after other features). We assume that although this feature is put first in the causal ordering, the causal \gls{shap} algorithm will uncover that this feature only has a direct effect on the prediction, and therefore assign it an indirect causal connection strength close to 0 to the following features in the causal ordering. 

Another weakness is that, according to our causal ordering, $q_1, q_2, q_3$ (the joint features), have a causal effect on $q_4$ (the gripper feature). This is not necessarily true, but the current implementation does not allow two features to affect a third feature without affecting each other in the causal ordering. These two issues are highlighted in~\Cref{fig:causal_graph},  indicating that $\theta_{target}$ does not influence features succeeding it in the causal ordering and that the joint variables do not influence $q_1$.

\section{Results and Discussion}\label{sec:results_discussion}
In this section, we discuss and compare the results from using causal SHAP and KernelSHAP to explain the actions of the agent performing the robotic lever manipulation task. To do this, we select three events from the dataset described above. The first of these events is from episode 1, where the manipulator pushes the lever from the start angle to the target angle with the gripper closed. We select a time-step in which the manipulator is actively pushing the manipulator and analyze it. This event is hereafter referred to as the \emph{pushing event}. The second event selected is from episode 3, in which the manipulator is grasping the lever before pulling it.  We here analyze the exact time-step in which the manipulator grasps the lever. We refer to this event as the \emph{grasping event}. The third and final event we analyze also belongs to episode 3 and takes place just after the grasping event when the manipulator is being used to pull the lever from the start angle to the target angle. This event is referred to as the \emph{pulling event}. 

\subsection{Pushing event} 
\begin{figure}
    \centering
    \includegraphics[width=\linewidth]{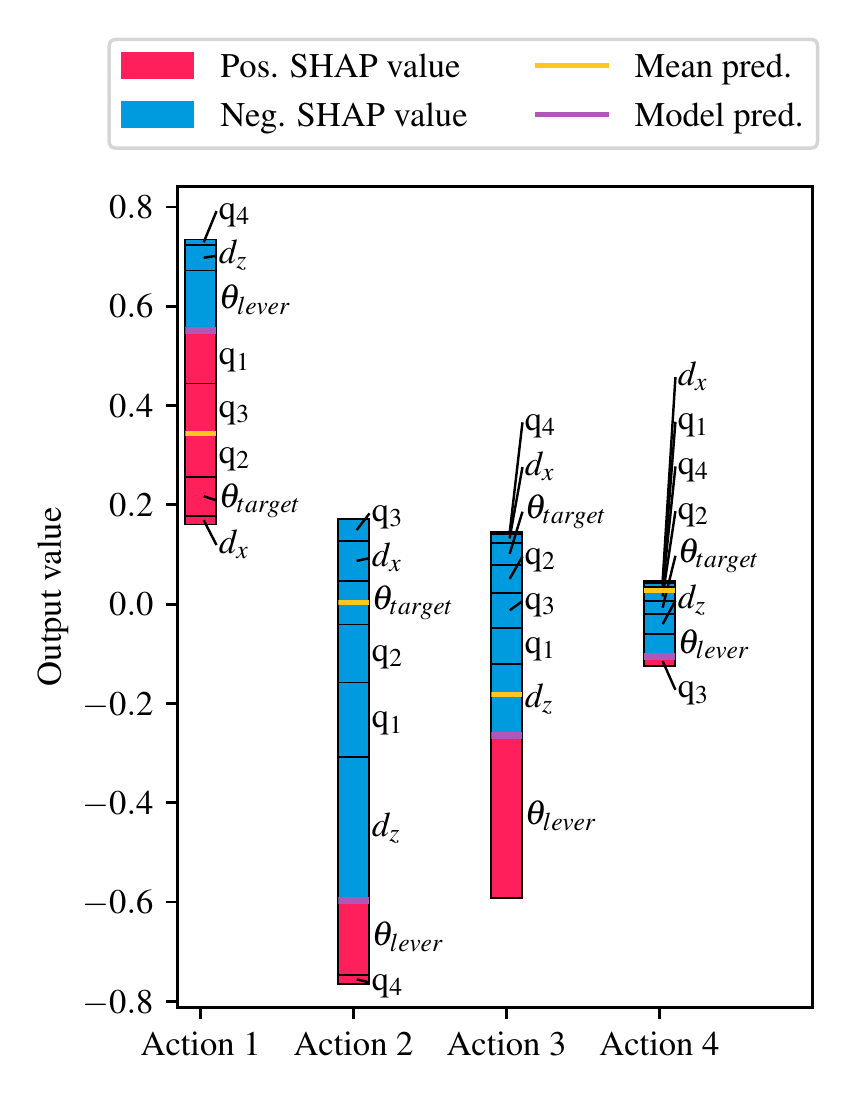}
    \caption{Episode 1: pushing event, causal SHAP}
    \label{fig:ep1_pushing_causal}
\end{figure}
\begin{figure}
    \centering
    \includegraphics[width=\linewidth]{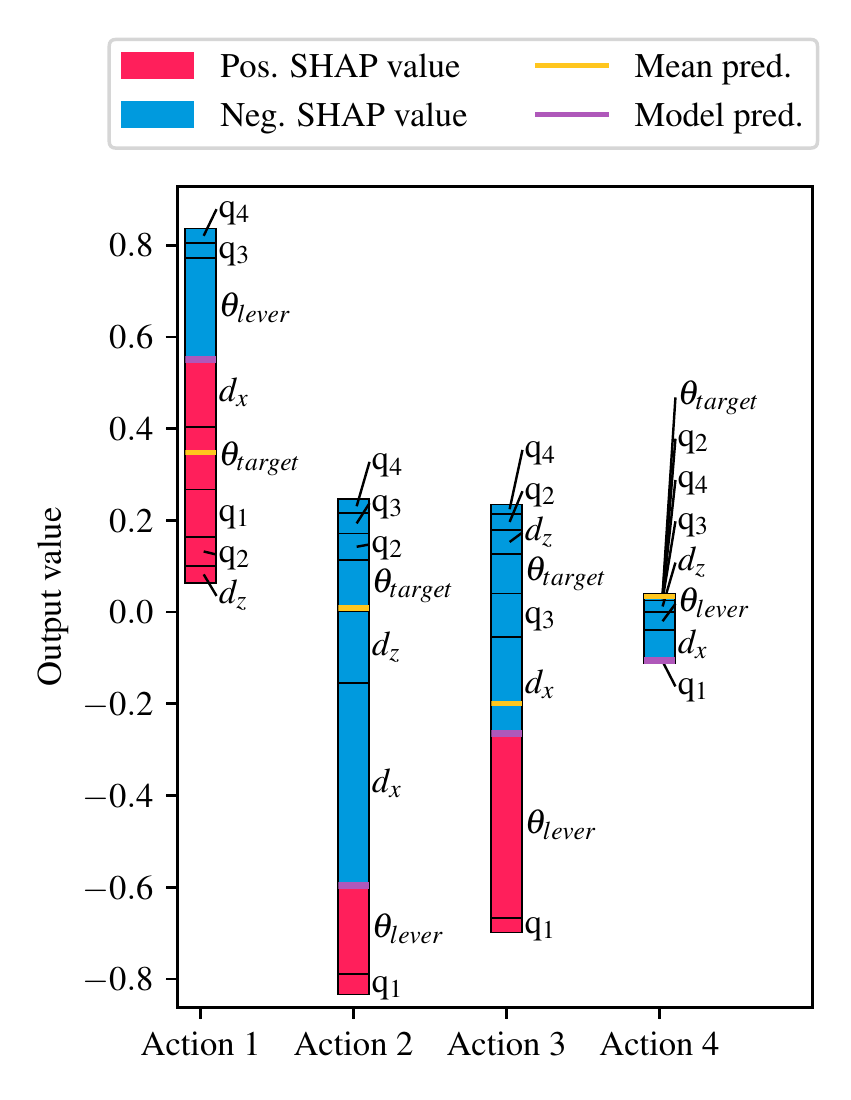}
    \caption{Episode 1: pushing event, KernelSHAP}
    \label{fig:ep1_pushing_kernel}
\end{figure}
For the pushing event, the force plots for causal \gls{shap} are shown in \Cref{fig:ep1_pushing_causal} and for KernelSHAP in \Cref{fig:ep1_pushing_kernel}. In both plots, for action 4, we can see that all but one feature have negative \gls{shap} values, which tells us that most aspects of this situation inform the agent to keep the gripper closed. For both methods, and for the actions corresponding to moving the joints (actions 1 to 3), we see that $q_4$ is the feature with the lowest \gls{shap} value. This means that both methods agree that it is not important whether the gripper is closed for the movement of the joints. Both methods assign similar \gls{shap} values to $\theta_{lever}$ and $\theta_{target}$. However, we can see that causal \gls{shap} assigns slightly lower values to $\theta_{lever}$ compared to KernelSHAP. This is likely because this feature is at the bottom of the causal ordering. Nevertheless, $\theta_{lever}$ is still among the features with the highest \gls{shap} values. This tells us that $\theta_{lever}$ is an important predictor variable for how the manipulator should be used to move the lever; in other words, it is important to know where the lever is in order to move it. 

In general, we see that causal \gls{shap} gives higher \gls{shap} values to the joint variables, $q_1, q_2, q_3$. In contrast, KernelSHAP gives higher values to $d_x$ and $d_z$, the features that together form a Cartesian vector from the end-effector to the lever's base. The joint variables are higher in the causal ordering, which is why causal \gls{shap} gives more importance to the joint variables. These two sets of features contain much of the same information, that is, information about the manipulator's position. However, in addition to this information about the manipulator's position, the joint variables contain information about the manipulator's orientation, while $d_x$ and $d_z$ contain information about where the lever is situated in relation to the manipulator. The exclusive information that these two sets of features contain makes each of them valuable for performing the lever control task. However, because the joint variables are, in fact, what is being controlled by the actions $1-3$, and it is difficult to control something one does not know where is, it stands to reason that these features should have among the highest \gls{shap} values.

\subsection{Grasping event} 

\begin{figure}
    \centering
    \includegraphics[width=\linewidth]{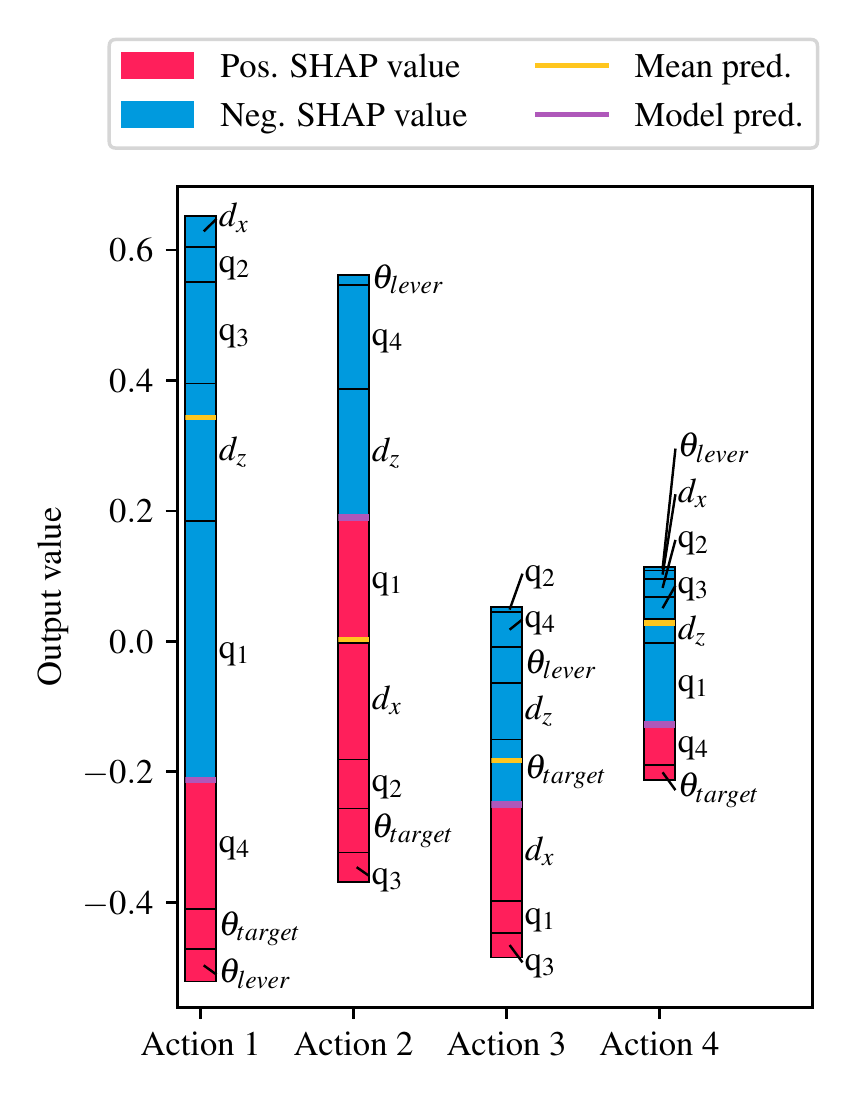}
    \caption{Episode 3: grasping event, causal SHAP}
    \label{fig:ep3_grasping_causal}
\end{figure}

\begin{figure}
    \centering
    \includegraphics[width=\linewidth]{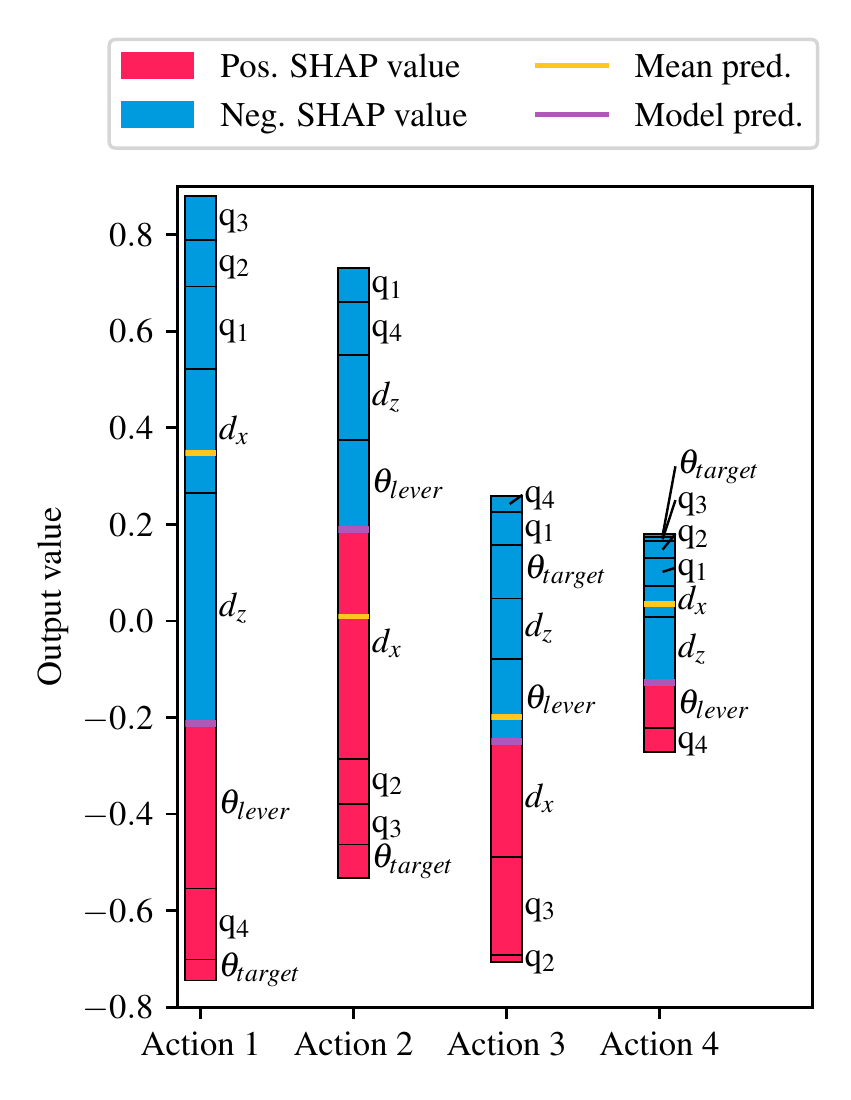}
    \caption{Episode 3: grasping event, KernelSHAP}
    \label{fig:ep3_grasping_kernel}
\end{figure}

For the grasping event, the plots for causal SHAP and KernelSHAP are shown in \Cref{fig:ep3_grasping_causal} and \Cref{fig:ep3_grasping_kernel}, respectively. In contrast to the plots for the pushing event, $q_4$ is not the feature with the lowest \gls{shap} value. This is presumably because the agent intends to pull the lever, and $q_4$ is important for knowing that the lever still needs to be grasped by the manipulator.

At this point, we see more significant differences between the two methods than we did during the pushing event, with regards to the magnitude of the $\theta_{lever}$ feature's \gls{shap} value. Again, KernelSHAP assigns the most importance to this feature. In fact, for causal SHAP, $\theta_{lever}$ overall has the lowest \gls{shap} value in the grasping event. This is curious since the position of the lever should be highly important for deciding how to grasp the lever. This is likely because causal \gls{shap} assigns some of the contribution from $\theta_{lever}$ to features above it in the causal ordering. 

Similar to the pushing event, the joint variables are generally more important according to causal SHAP than they are according to KernelSHAP. Conversely, $d_x$ and $d_z$ are generally more important according to KernelSHAP than what they are to causal SHAP. However, there are some exceptions to this: Consider $q_2$, which has approximately the same \gls{shap} values for both plots' corresponding actions, and action $3$, for which the \gls{shap} value of $q_3$ has a larger magnitude from KernelSHAP than from causal SHAP.

\subsection{Pulling event} 

\begin{figure}
    \centering
    \includegraphics[width=\linewidth]{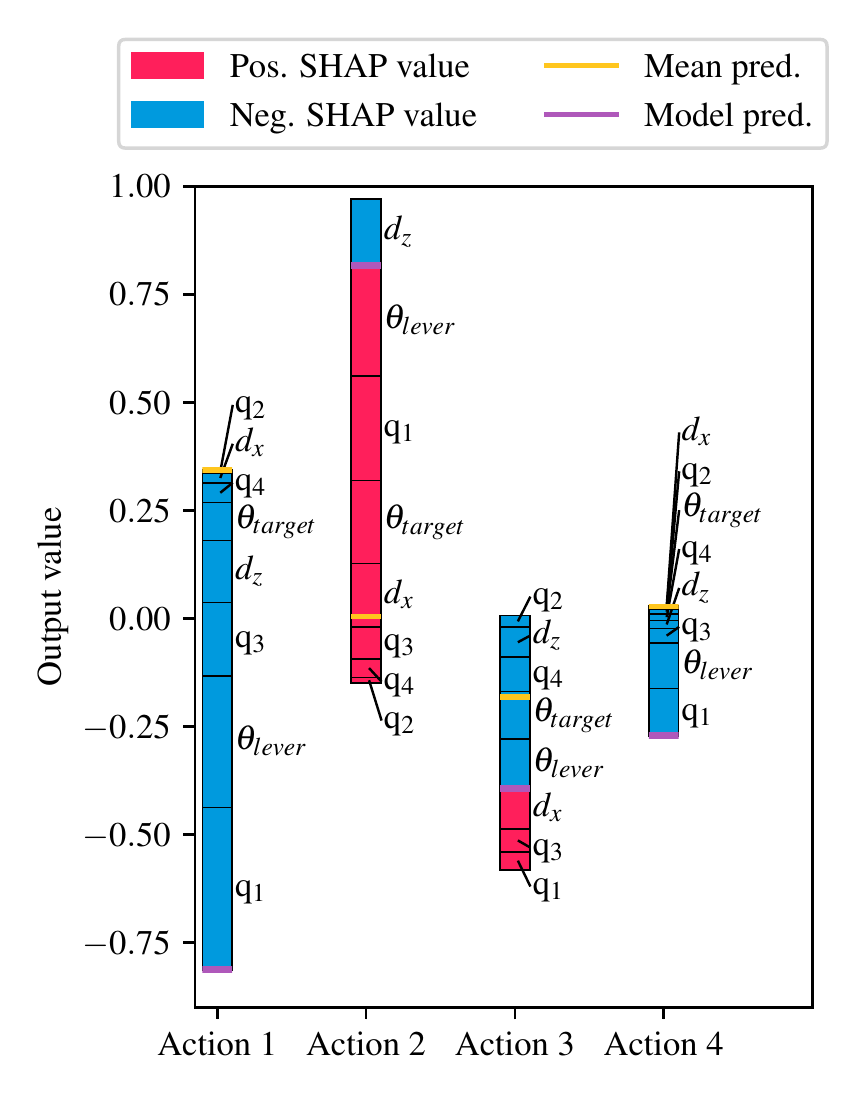}
    \caption{Episode 3: pulling event, causal SHAP}
    \label{fig:ep3_pulling_causal}
\end{figure}

\begin{figure}
    \centering
    \includegraphics[width=\linewidth]{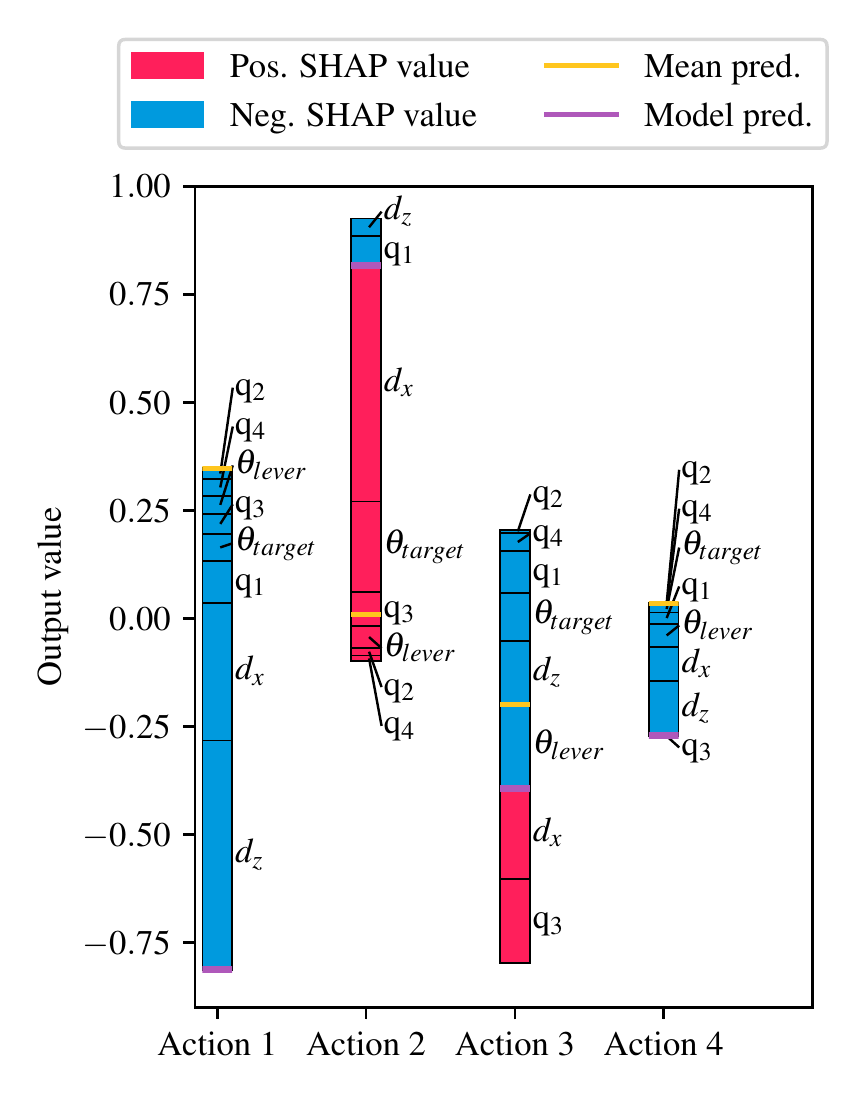}
    \caption{Episode 3: pulling event, KernelSHAP}
    \label{fig:ep3_pulling_kernel}
\end{figure}

\Cref{fig:ep3_pulling_causal} and \Cref{fig:ep3_pulling_kernel} show the results from causal SHAP and KernelSHAP, respectively. Here, we see even more clearly what we have seen in the two previous events regarding KernelSHAP prioritizing $d_x$ and $d_z$ over the joint variables, and vice versa for causal SHAP. In fact, $d_x$ and $d_z$ account for over half of the total magnitude of all the \gls{shap} values for actions $1$, $2$ and $4$ in \Cref{fig:ep3_pulling_kernel}.

As discussed in \Cref{sec:methodology_xai}, $\theta_{target}$ was put on the top of the causal ordering by necessity. However, we can see that this feature has approximately the same \gls{shap} value for all actions in the pulling event for both of the methods. This was also the case for the pushing event and the grasping event described above. This suggests that the causal \gls{shap} algorithm has discovered that this feature only has a direct effect, and therefore has given an indirect causal connection strength of approximately $0$ to the features succeeding it in the causal graph, in agreement with our expectation. 

\section{Conclusion}\label{sec:conclusion}
We have shown how causal \gls{shap} can be used to explain not only the direct effects but also the indirect effects a feature can have on the decisions of a \gls{drl} agent controlling a real-world robotics system. In addition, we have shown how causal \gls{shap} allows for the incorporation of domain knowledge in the explanation generation process. We expect that the causal relations will be an essential part of \gls{xai} going forward, especially for explaining models of physical systems. 

Because of the complex nature of the explanations shown in this paper, we recognize that these are most useful for data scientists, model developers, and others with experience in data analysis and \gls{xai}. When it comes to explanations for end-users that are non-experts, our explanations would likely need to be processed. According to \cite{miller_explanation_2019}, explanations should be contrastive, and explanations based on \emph{counterfactuals} \cite{hume2000enquiry} are more intuitive to humans. Work has been done towards unifying feature attribution and counterfactuals in \cite{Kommiya_Mothilal_2021}, and also towards generating counterfactuals from \gls{shap}, as in \cite{rathi2019generating}. Further work could, therefore, consist of researching whether transforming these feature attribution explanations to counterfactual explanations could make them more convenient for non-experts. 

More specific to causal \gls{shap}, further work can, among others, consist in altering the implementation of causal \gls{shap} to account for fully independent features (such as the $\theta_{target}$ in this paper). Another valuable addition to the implementation would be the possibility to specify manually which indirect causal connections should have strength $0$ (corresponding to causally independent subsystems). 

\section*{Acknowledgment}
The Research Council of Norway supported this work through the EXAIGON project, project number 304843.

\bibliographystyle{IEEEtran}
\bibliography{mylib}

\end{document}